%% file: main.tex
\pdfoutput=1
\documentclass[conference]{IEEEtran}
\IEEEoverridecommandlockouts
\usepackage{cite}
\usepackage{amsmath,amssymb,amsfonts}
\usepackage{algorithmic}
\usepackage{graphicx}
\usepackage{textcomp}
\usepackage{booktabs}
\usepackage[table]{xcolor}
\usepackage{url}
\usepackage{siunitx}
\usepackage{hyperref}
\DeclareMathOperator*{\argmin}{argmin}
\usepackage{framed}

\makeatletter
\let\MYcaption\@makecaption
\makeatother
\usepackage[font=footnotesize]{subcaption}
\makeatletter
\let\@makecaption\MYcaption
\makeatother

\definecolor{lightgray}{rgb}{0.83, 0.83, 0.83}

\def\BibTeX{{\rm B\kern-.05em{\sc i\kern-.025em b}\kern-.08em
    T\kern-.1667em\lower.7ex\hbox{E}\kern-.125emX}}
\begin{document}

\title{Comparing Offline and Online Testing of Deep Neural Networks: An Autonomous Car Case Study
}

\author{
\IEEEauthorblockN{
Fitash Ul Haq\IEEEauthorrefmark{1}, 
Donghwan Shin\IEEEauthorrefmark{1}, 
Shiva Nejati\IEEEauthorrefmark{1},
and Lionel Briand\IEEEauthorrefmark{1}\IEEEauthorrefmark{2}}

\IEEEauthorblockA{\IEEEauthorrefmark{1}University of Luxembourg, Luxembourg\\
Email: \{fitash, donghwan, nejati, briand\}@svv.lu}
\IEEEauthorblockA{\IEEEauthorrefmark{2}University of Ottawa, Canada\\
Email: lbriand@uOttawa.ca}
}

\maketitle

\begin{abstract}
There is a growing body of research on developing testing techniques for Deep Neural Networks (DNN). We distinguish two general modes of testing for DNNs: \emph{Offline} testing where DNNs are tested as individual units based on test datasets obtained independently from the DNNs under test, and \emph{online} testing where DNNs are embedded into a specific application and tested in a close-loop mode in interaction with the application environment. In addition, we identify two sources for generating test datasets for DNNs: Datasets obtained from \emph{real-life} and datasets generated by \emph{simulators}.  While offline testing can be used with datasets obtained from either sources, online testing is largely confined to using simulators since online testing within real-life applications can be time consuming, expensive and dangerous. In this paper, we study the following two important questions aiming to compare test datasets and testing modes for DNNs: \emph{First, can we use simulator-generated data as a reliable substitute to real-world data for the purpose of DNN testing?} \emph{Second, how do online and offline testing results differ and complement each other?} Though these questions are generally relevant to all autonomous systems, we study them in the context of automated driving systems where, as study subjects, we use DNNs automating end-to-end control of cars' steering actuators. Our results show that simulator-generated datasets are able to yield DNN prediction errors that are similar to those obtained by testing DNNs with real-life datasets. Further, offline testing is more optimistic than online testing as many safety violations identified by online testing could not be identified by offline testing, while large prediction errors generated by offline testing always led to severe safety violations detectable by online testing.
\end{abstract}


\input{intro}
\input{background}
\input{experiment}
\input{related-work}

\input{conclusion}

\section*{Acknowledgment}
We gratefully acknowledge funding from the  European Research Council (ERC) under the European Union's Horizon 2020 research and innovation programme (grant agreement No 694277) and from IEE S.A. Luxembourg.

\bibliographystyle{IEEEtran}
\bibliography{DNN-ADAS-Testing}

\end{document}

%% file: intro.tex
\section{Introduction}
\label{sec:intro}
Deep Neural Networks 
(DNN)~\cite{chen2015deepdriving,bojarski2016end,chi2017deep} have made 
unprecedented progress largely fueled by increasing availability of data and 
computing powers. DNNs have been able to automate challenging real-world tasks 
such as image classification~\cite{ciresan2012}, natural language 
processing~\cite{SutskeverVL14} and speech recognition~\cite{DengHK13}, making 
them key enablers of smart and autonomous systems such as automated-driving 
vehicles. As DNNs are increasingly used in safety critical autonomous systems, 
the challenge of ensuring safety and reliability of DNN-based systems emerges as 
a difficult and fundamental software verification problem. 

Many DNN testing approaches have been proposed 
recently~\cite{DeepXplore,DeepTest,DeepRoad,DeepGauge,zhou2018deepbillboard}. 
Among these, we distinguish two high-level, distinct approaches to DNN testing: 
(1)~Testing DNNs as stand-alone components, and (2)~testing DNNs embedded into a 
specific application (e.g., an automated-driving system) and in interaction with 
the application environment. We refer to the former as \emph{offline testing} 
and to the latter as \emph{online testing}. Specifically, in offline testing, 
DNNs are tested as a unit in an open-loop mode. They are fed with test inputs 
generated independently from the DNN under test, either manually or 
automatically (e.g., using image generative methods~\cite{DeepRoad}). The 
outputs of DNNs are then typically evaluated by assessing their prediction 
error, which is the difference between the expected test outputs (i.e., test 
oracles) and the outputs generated by the DNN under test. In online testing, 
however, DNNs are tested within an application environment in a closed-loop 
mode. They receive test inputs generated by the environment, and their outputs 
are, then, directly fed back into the environment. Online testing evaluates  
DNNs by monitoring the requirements violations they trigger, for example related 
to safety.  

There have been several offline and online DNN testing approaches in the 
literature~\cite{zhang2019machine}. However, comparatively, offline testing has 
been far more studied to date. This is partly because offline testing does not 
require the DNN to be embedded into an application environment and can be 
readily carried out with either manually generated or automatically generated 
test data. Given the increasing availability of open-source data, a large part 
of offline testing research uses open-source, manually-generated real-life test 
data. Online testing, on the other hand, necessitates embedding a DNN into an 
application environment, either real or simulated. Given the safety critical 
nature of many systems relying on DNN (e.g., self-driving cars), most online 
testing approaches rely on simulators, as testing DNNs embedded into real and 
operational environment is expensive, time consuming and can be dangerous in 
some cases. 

While both offline and online testing approaches have shown to be promising, 
there is limited insight as to how these two approaches compare with one 
another. While, at a high-level, we expect offline testing to be faster and less 
expensive than online testing, we do not know how they compare with respect to 
their ability to reveal faults, for example leading to safety violations. 
Further, we would like to know if large prediction errors identified by offline 
testing always lead to safety violations detectable by online testing? or if the 
safety violations identified by online testing translate into large prediction 
errors? Answers to these questions enable us to better know the relationships 
and the limitations of the two testing approaches. We can then know which 
approach is to be recommended in practice for testing autonomous systems, or if 
the two are somehow complementary and should be best combined.

In this paper, though the investigated questions are generally relevant to all 
autonomous systems, we perform an empirical study to compare DNN offline and 
online testing in the context of automated driving systems. In particular, our 
study aims to ultimately answer the following research question: \emph{RQ1: How 
do offline and online testing results differ and complement each other?} To 
answer this question, we use open-source DNN models developed to automate 
steering functions of self-driving vehicles~\cite{udacity:challenge}. To enable 
online testing of these DNNs, we integrate them into a powerful, high-fidelity 
physics-based simulator of self-driving cars~\cite{prescan}. The simulator 
allows us to specify and execute scenarios capturing various road traffic 
situations, different pedestrian-to-vehicle and vehicle-to-vehicle interactions, 
and different  road-topologies, weather conditions and infrastructures. As a 
result, in our study offline and online testing approaches are compared with 
respect to the data generated automatically using a simulator. To ensure that 
this aspect does not impact the validity of our comparison, we investigate the 
following research question as a pre-requisite of the above question: \emph{RQ0: 
Can we use simulator-generated data as a reliable substitute to real-world data 
for the purpose of DNN testing?}

To summarize, the main contribution of this paper is that we provide, for the 
first time, an empirical study comparing offline and online testing of DNNs. Our 
study investigates two research questions RQ0 and RQ1 (described above) in the 
context of an automated-driving system. Specifically, \begin{enumerate}
\item RQ0: Our results show that simulator-generated datasets are able to yield 
DNN prediction errors that are similar to those obtained by testing DNNs with 
real-life datasets. Hence, simulator-generated data can be used in lieu of 
real-life datasets for testing DNNs in our application context.
\item RQ1: We found that offline testing is more optimistic than online testing 
because the accumulation of prediction errors over time is not observed in 
offline testing. Specifically, many safety violations identified by online 
testing could not be identified by offline testing as they did not cause large 
prediction errors. However, all the large prediction errors generated by offline 
testing led to severe safety violations detectable by online testing.
\end{enumerate}

To facilitate the replication of our study, we have made all the experimental 
materials, including simulator-generated data, publicly available~\cite{supp}.

The rest of the paper is organized as follows. Section~\ref{sec:background} 
provides background on DNNs for autonomous vehicles, introduces offline and 
online testing, describes our proposed domain model that is used to configure 
simulation scenarios for automated driving systems, and formalizes the main 
concepts in offline and online testing used in our experiments. 
Section~\ref{sec:expr} reports the empirical evaluation. 
Section~\ref{sec:offline-online} surveys the existing research on online and 
offline testing for automated driving system. Section~\ref{sec:conclusion} 
concludes the paper.

%% file: background.tex
\section{Offline and Online Testing frameworks}\label{sec:background}
This section provides the basic concepts that will be used throughout the paper.

\begin{figure}
	\centering
	\includegraphics[width=\linewidth]{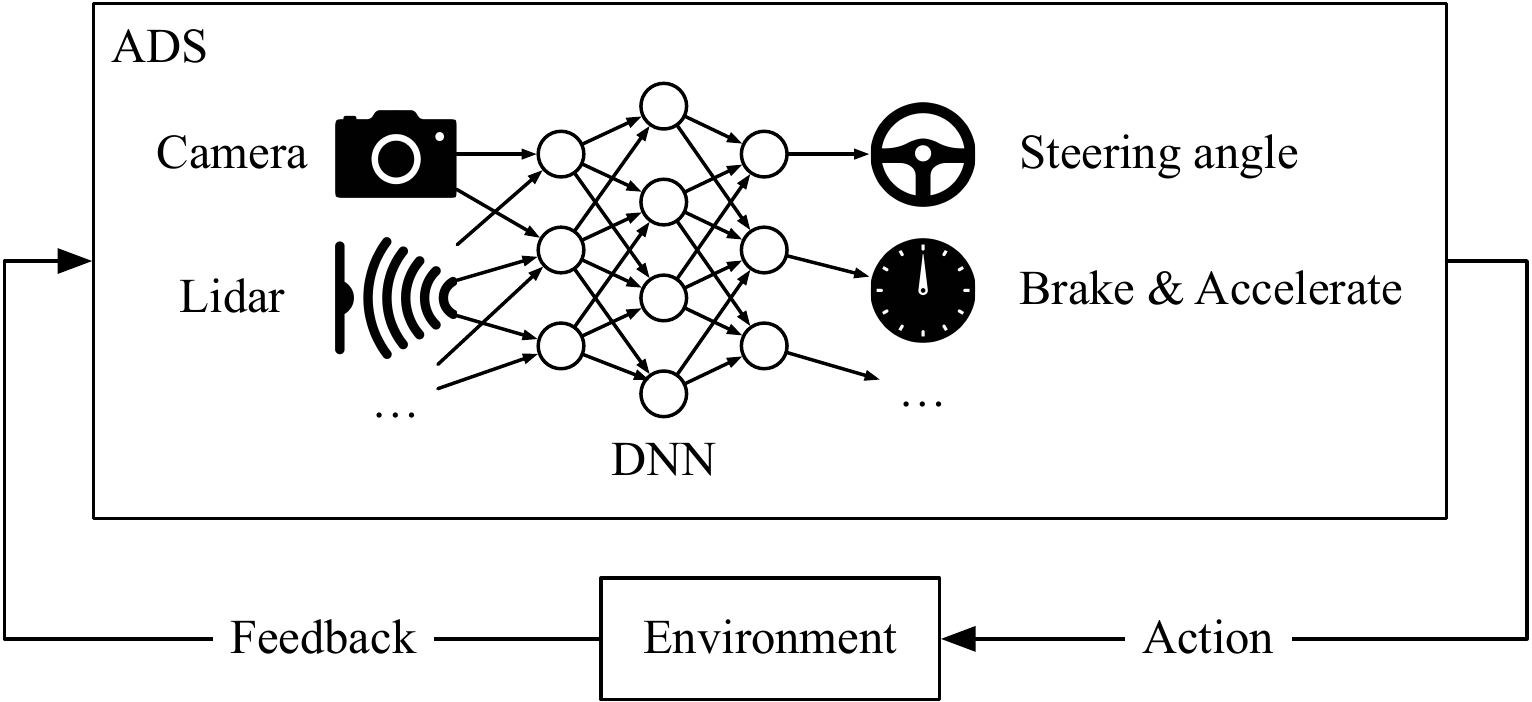}
	\caption{Overview of DNN-based ADS}
	\label{fig:dnn-in-ads}
\end{figure}

\subsection{DNNs in ADS}
Depending on the ADS design, DNNs may be used in two ways to automate the 
driving task of a vehicle: One design approach is to incorporate DNNs into the 
perception layer of ADS primarily to do \emph{semantic 
segmentation}~\cite{6248074}, i.e., to classify and label each and every pixel 
in a given image. The software controller of ADS then decides what commands 
should be issued to the vehicle's actuators based on the classification results 
produced by the DNN~\cite{pomerleau1989alvinn}. An alternative design approach 
is to use DNNs to perform the \emph{end-to-end} control of a 
vehicle~\cite{udacity:challenge} (e.g., Figure~\ref{fig:dnn-in-ads}). In this 
case, DNNs directly generate the commands to be sent to the vehicles' actuators 
after processing images received from cameras. Our approach to compare offline 
and online testing of DNNs for ADS is applicable to both ADS designs. In the 
comparison provided in this paper, however, we use DNN models automating the 
end-to-end control of steering function of ADS since these models are publicly available 
online and have been extensively used in recent papers on DNN 
testing~\cite{DeepTest,DeepRoad,DeepGauge,Surprise}. 
In particular, we use the DNN models from the Udacity self-driving challenge as our study 
subjects~\cite{udacity:challenge}. We refer to this class of DNNs as ADS-DNNs in the remainder 
of the paper. Specifically, ADS-DNN receives inputs from a camera,
and generates a steering angle command. 

\begin{figure}
	\centering
	\includegraphics[width=1\linewidth]{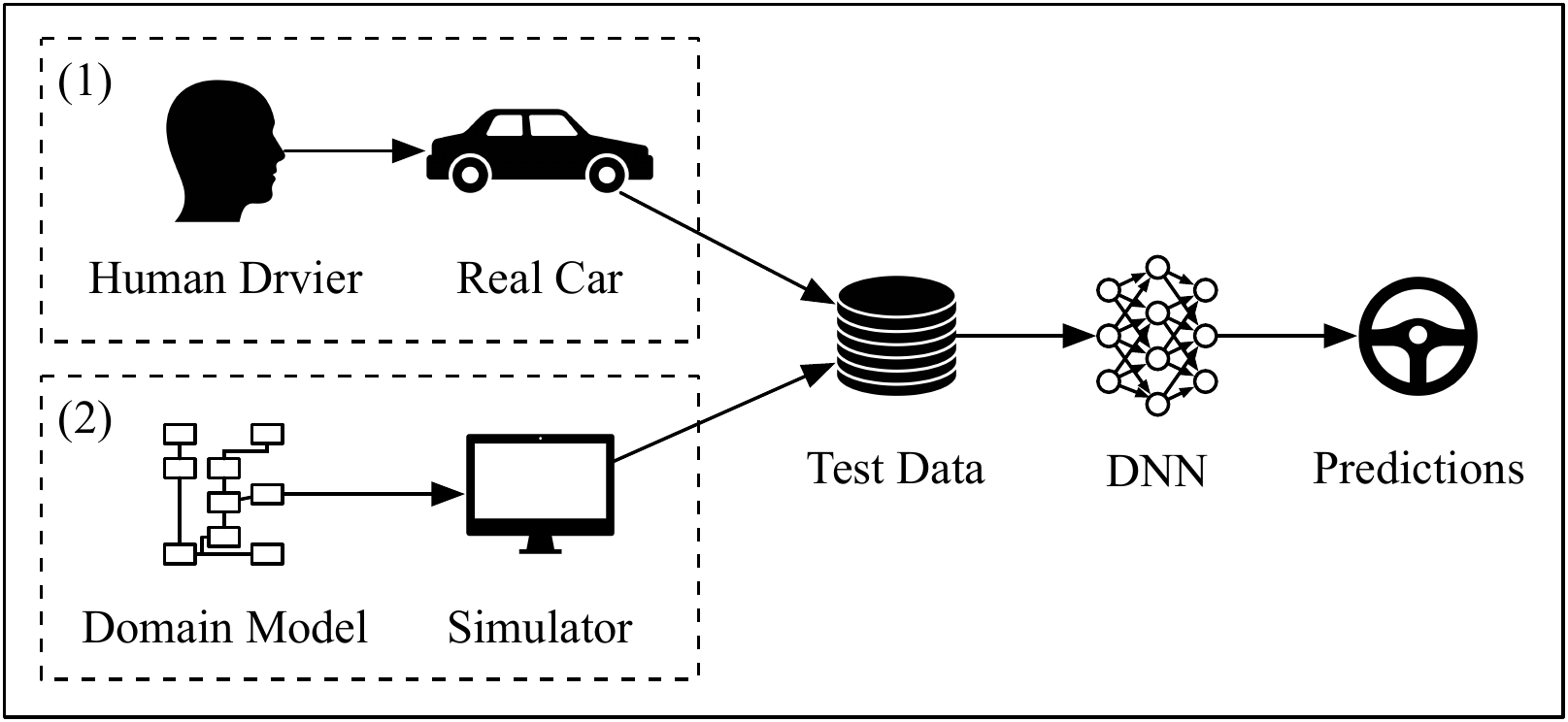}
	\caption{Offline testing using (1) real-world data and (2) simulator-generated data}
	\label{fig:offline}
\end{figure}

\subsection{Offline Testing}
\label{subsec:offline}
Figure~\ref{fig:offline} represents an overview of offline testing of DNN in the 
context of ADS. In general, a dataset used to test a DNN (or any ML model for 
that matter) is expected to be realistic to be able to provide an unbiased 
evaluation of the DNN under test. As shown in Figure~\ref{fig:offline}, we 
identify two sources for generating test data for the offline mode: (1) 
datasets captured from real-life driving, and (2) datasets generated by 
simulators. For our ADS-DNN models, a \emph{real-life dataset} is a video or a 
sequence of images captured by a camera mounted on a (ego) vehicle's dashboard 
while the vehicle is being driven by a human driver. The steering angle of the 
vehicle applied by the human driver is recorded for the duration of the video and 
each image (frame) of the video in this sequence is labelled by its corresponding steering 
angle. This yields a sequence of manually labelled images to be used for 
testing DNNs. There are, however, some drawbacks with test datasets captured 
from real-life. Specifically, data generation is expensive, time consuming and 
lacks diversity. The latter issue is particularly critical since driving scenes, 
driving habits, as well as objects, infrastructures and roads in driving scenes, 
can vary widely across countries, continents, climates, seasons, day times, and 
even drivers. 

As shown in Figure~\ref{fig:offline}, another source of test data generation 
for DNN offline testing is to use simulators to automatically generate videos 
capturing various driving scenarios. There are increasingly more high-fidelity 
and advanced physics-based simulators for self-driving vehicles fostered by the 
needs of the automotive industry which increasingly relies on simulators to 
improve their testing and verification practices. There are several examples of 
commercial ADS simulators (e.g., PreScan~\cite{prescan} and Pro-SiVIC~\cite{prosivic}) 
and a number of open source ones (e.g., CARLA~\cite{carla} and Apollo~\cite{apollo}). 
These simulators incorporate dynamic models of vehicles (including vehicles' actuators, 
sensors and cameras) and humans as well as various environment aspects (e.g., 
weather conditions, different road types, different infrastructures). The 
simulators are highly configurable and can be used to generate desired driving 
scenarios. In our work, we use the PreScan simulator to generate 
test datas for ADS-DNNs. PreScan is a widely-used, high-fidelity commercial ADS 
simulator in the automotive domain and has been used by our industrial partner.
In Section~\ref{sec:domain}, we present the domain model we define 
to configure the simulator, and describe how we automatically generate scenarios 
that can be used to test ADS-DNNs. Similar to real-life videos, the videos 
generated by our simulator are sequences of labelled images such that each image is 
labelled by a steering angle. In contrast to real-life videos, the steering 
angles generated by the simulator are automatically computed based on the road 
trajectory as opposed to being generated by a human driver. 

The simulator-generated test datasets are cheaper and faster to produce compared 
to real-life ones. In addition, depending on how advanced and comprehensive the 
simulator is, we can achieve a higher-level of diversity in the 
simulator-generated datasets by controlling and varying the objects, roads, 
weather, day time and infrastructures and their various features. However, it is 
not yet clear whether simulator-generated images can be used in lieu of real 
images since real images may have higher resolution, showing more natural 
texture and look more realistic. In this paper, we conduct an empirical study in 
Section~\ref{sec:expr} to investigate \emph{if we can use simulator-generated 
images as a reliable alternative to real images for testing ADS-DNNs}.

\subsection{Online Testing}
\label{subsec:online}

\begin{figure}
	\centering
	\includegraphics[width=	\linewidth]{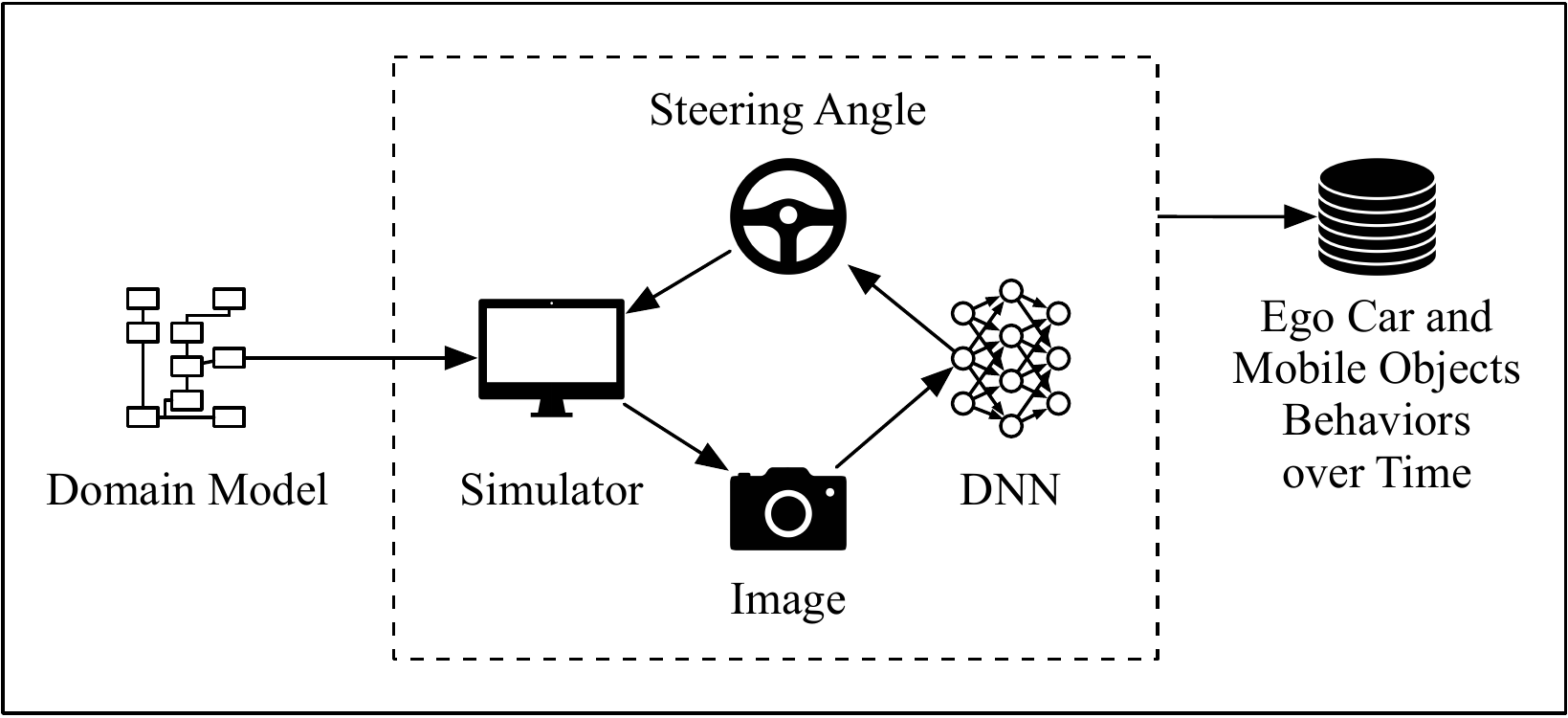}
	\caption{Online testing of ADS-DNNs using simulators}
	\label{fig:online}
\end{figure}

Figure~\ref{fig:online} provides an overview of online testing of DNNs in the 
context of ADS. In contrast to offline testing, DNNs are embedded into a 
simulator, they receive images generated by the simulator, and their outputs 
are directly sent to the (ego) vehicle models of the simulator. In this paper, we  
embed the ADS-DNN into PreScan by providing the former with the outputs from the 
camera model in input and connecting the steering angle output of the ADS-DNN as 
input command to the vehicle dynamic model. With online testing, we can evaluate how 
predictions generated by an ADS-DNN for an image generated at time $t$ in a 
scenario impacts the images to be generated at the time steps after $t$. 
Specifically, if the ADS-DNN orders the ego vehicle to turn with an angle $\theta$ at 
time $t$ during a simulation, the camera's field of view will be shifted by 
$\theta$ within a small time duration $t_d$, and hence, the image captured at 
time $t+t_d$ will account for the modified camera's field of view. 
Note that $t_d$ is the time required by the vehicle to actually perform a 
command and is computed by the dynamic model in the simulator. With online 
testing, in addition to the steering angle outputs directly generated by the 
ADS-DNN, we obtain the trajectory outputs of the ego vehicle which enable us to 
determine whether the car is able to stay in its lane. 

Note that one could perform online testing with a real car and collect real-life 
data. However, this is expensive, very dangerous, in particular for end-to-end 
DNNs such as ADS-DNN, and can only be done under very restricted conditions on 
some specific public roads. 

We conduct an empirical study in Section~\ref{sec:expr} to investigate \emph{How 
offline and online testing results differ and complement each other for 
ADS-DNNs}.

\subsection{Domain Model}\label{sec:domain}
\begin{figure}
	\centering
	\includegraphics[width=.9\linewidth]{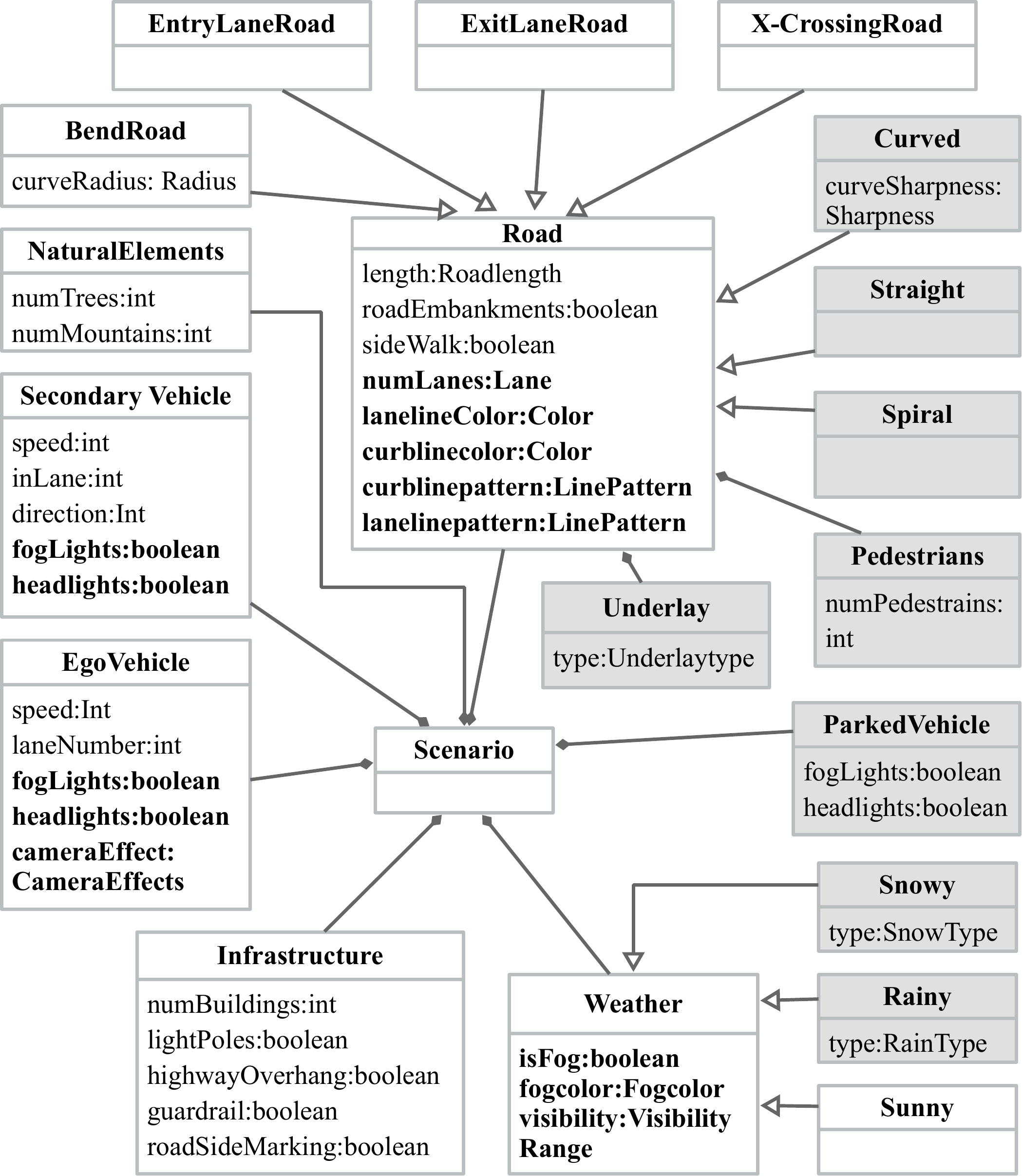}
	\caption{Partial domain model for scenario generation}
	\label{fig:dm}
\end{figure}

Figure~\ref{fig:dm} shows a fragment of the domain model capturing the test input 
space of ADS-DNN. To develop the domain model, we relied on the features that we 
observed in the real-world test datasets for ADS-DNN  (i.e., the Udacity testing 
datasets~\cite{udacity:dataset}) 
as well as the configurable features of our simulator. The domain model 
includes different types of road topologies (e.g., straight, curved, with entry 
or exit lane), different weather conditions (e.g., sunny, foggy, rainy, snowy), 
infrastructure (e.g., buildings and overhead hangings), nature elements (e.g., 
trees and mountains), an ego vehicle, secondary vehicles and pedestrians. 
Each entity has multiple variables. For example, an ego vehicle has the following 
variables: a speed, a number (id) identifying the lane in which it is driving, 
a Boolean variable indicating if its fog lights are on or off, and many others. 
In addition to entities and variables, our domain model includes some 
constraints describing valid value assignments to the variables. These 
constraints mostly capture the physical limitations and traffic rules. For 
example, the vehicle speed cannot be higher than some limit on steep curved roads.  
We have specified these constraints in the Object Constraint Language 
(OCL)~\cite{OCL}. The complete domain model, together with the OCL constraints, 
are available in the supporting materials~\cite{supp}. 

To produce a simulation scenario (or test scenario) for ADS-DNN, we develop an 
\emph{initial configuration} based on our domain model. 
An initial configuration is a vector of values assigned to the variables in the 
domain model and satisfying the OCL constraints. The simulator generates for 
each of the mobile objects defined in a scenario, namely the ego vehicle and 
secondary vehicles and pedestrians, a vector of the trajectory path of that 
object (i.e., a vector of values indicating the positions and speeds of the 
mobile object). The length of the vector is determined by the duration of the 
simulation. The position values are computed based the characteristics of 
the static objects, specified by the initial configuration such as roads and 
sidewalks, as well as the speed of the mobile objects.

\subsection{Formalization}

\begin{table}
\centering
\caption{Comparison between offline and online testing}
\label{table:online-offline}
\renewcommand{\arraystretch}{1.3}
\rowcolors{1}{white}{lightgray}
\begin{tabular}{l|p{2.8cm}p{2.8cm}}
\toprule
Criterion & Offline testing & Online testing  \\
\midrule
%\midrule
Definition		
& Test DNNs using historical data already generated manually or automatically
& Test DNNs by embedding them into an application environment (virtual or real) \\
%\midrule
Test mode
&	Open-loop
& Closed-loop	\\
%\midrule
Test input
& Sequences of images from a camera (real-life) or a camera model (in a simulator)
& An initial configuration for a simulator to guide the simulator into generating a specific driving scenario	\\
%\midrule
Test output
& Prediction error
& Safety violation \\
%\midrule
Execution time & Low & High \\
\bottomrule
\end{tabular}
\end{table}

Table~\ref{table:online-offline} summarizes the comparison between offline and 
online testing as detailed in Sections~\ref{subsec:offline} and 
\ref{subsec:online}. Briefly, offline testing verifies the DNN using historical 
data consisting of sequences of images captured from real-life camera or based 
on a camera model of a simulator. 
In either case, the images are labelled with the steering angles. Offline testing 
measures the \emph{prediction errors} of the DNN to evaluate test results. In 
contrast, online testing verifies the DNN embedded into an application 
environment in a closed-loop mode. The test inputs for online testing are 
initial configurations of the simulator, generated based on our domain model (see 
Section~\ref{sec:domain}), that guide the generation of specific 
scenarios. The output of online testing is whether, or not, for a given 
simulation scenario, a safety violation has happened. In our context, a safety 
violation happens when the ego car strays out of its lane such that it may risk an 
accident. 
Since offline testing relies on historical data, it has a low execution time.  
However, the time required to perform online testing is relatively 
high because it encompasses the time required for the DNN-based ADS to execute 
and interact with its environment. Note that the execution time in 
Table~\ref{table:online-offline} only refers to the time required to perform 
testing and not the time or cost of generating test inputs. 

In the remainder of this section, we formalize inputs and outputs for offline and online testing.  
We denote a real-life test dataset by a sequence $\mathbf{r} = \langle (i^r_1, 
\theta^r_1), (i^r_2, \theta^r_2), \dots, (i^r_n, \theta^r_n) \rangle$ of tuples. 
For $j=1,\dots,n$, each tuple $(i^r_j, \theta^r_j)$ of $\mathbf{r}$ consists of an image $i^r_j$ 
and a steering angle $\theta^r_j$ label. A DNN $d$, when provided with a 
sequence $\langle i^r_1, i^r_2, \dots, i^r_n \rangle$ of the images of $\mathbf{r}$, 
returns a sequence $\langle \hat{\theta}^r_1, 
\hat{\theta}^r_2, \dots, \hat{\theta}^r_n \rangle$ of predicted steering angles. 
The prediction error of $d$ for $\mathbf{r}$ is, then, computed using two 
well-known metrics, Mean Absolute Error (MAE) and Root Mean Square Error (RMSE), 
defined below:

\begin{align*}
\mathit{MAE}(d, \mathbf{r}) &= \frac{\sum_{i=1}^{n} |\theta^r_i - \hat{\theta^r_i}|}{n}
\\
\mathit{RMSE}(d, \mathbf{r}) &= \sqrt{\frac{\sum_{i=1}^{n} (\theta^r_i - \hat{\theta^r_i})^2}{n}}
\end{align*}

To generate a test dataset using a simulator, we provide the simulator with an 
initial configuration of a scenario as defined in Section~\ref{sec:domain}. 
We denote the test dataset generated by a simulator for a scenario $s$ for offline 
testing by $\mathit{sim}(s) =
 \langle (i^s_1, \theta^s_1), (i^s_2, \theta^s_2), \dots, (i^s_n, \theta^s_n) \rangle$. 

For online testing, we embed a DNN $d$ into a simulator and run the simulator. 
For each (initial configuration of a) scenario, we execute the simulator for a time duration 
$T$. The simulator generates outputs as well as images at regular time steps 
$t_\delta$, generating outputs as vectors of size $m=\lfloor \frac{T}{t_\delta} 
\rfloor$. Each simulator output and image takes an index between $1$ to $m$. We 
refer to the indices as simulation time steps. At each time step $j$, the 
simulator generates an image $i^s_j$ to be sent to $d$ as input, and 
$d$ generates a predicted steering angle $\hat{\theta}^s_j$ which is sent to 
the simulator. The status of the ego car is then 
updated in the next time step $j+1$ (i.e., the time duration it takes to update 
the car is $t_\delta$) before the next image $i^s_{j+1}$ is generated. 
In addition to images, the simulator generates the position of the ego car over 
time. Since our DNN is for an automated lane keeping function, we use the 
Maximum Distance from Center of Lane (MDCL) metric for the ego car to determine 
if a safety violation has occurred. The value of MDCL is computed at the end of 
the simulation when we have the position vector of the ego car over time steps, which was 
guided by our DNN. 
We cap the value of MDCL at \SI{1.5}{\meter}, indicating that when MDCL 
is \SI{1.5}{\meter} or larger, the ego car has already departed its lane and a 
safety violation has occurred. To have a range between 0 and 1, MDCL is 
normalized by dividing 1.5 by the actual distance in meter.

%% file: experiment.tex
\section{Experiments}\label{sec:expr}
In this section, we compare offline and online testing of DNNs by answering the 
two research questions we have already motivated in Sections~\ref{sec:intro} 
and \ref{sec:background}, which are re-stated below:

\textbf{RQ0:} \emph{Can we use simulator-generated data as a reliable 
alternative source to real-world data?}
Recall that in Figure~\ref{fig:offline}, we described two sources for generating 
test data for offline testing.
As discussed there, simulator-generated test data is cheaper and faster to 
produce and is more amenable to input diversification compared to real-life test data. 
On the other hand, the texture and resolution of real-life data look more natural 
and realistic compared to the simulator-generated data. With RQ0, we aim to 
investigate whether, or not, such differences lead to 
significant inaccuracies in predictions of the DNN under test. To do so, we 
configure the simulator to generate a dataset (i.e., a sequence of labelled images) that 
closely resembles the characteristics of a given real-life dataset.  We then 
compare the offline testing results for these datasets. The answer to this 
question, which serves as a pre-requisite of our next question, will determine 
if we can rely on simulator-generated data for testing DNNs in either offline or 
online testing modes.

\textbf{RQ1:} \emph{How do offline and online testing results differ and 
complement each other?}
RQ1 is the main research question we want to answer in this paper.
It is important to know how the results obtained by testing a DNN irrespective of a 
particular application compare with test results obtained by embedding a DNN into 
a specific application environment.
The answer will guide engineers and researchers to better understand
the applications and limitations of each testing mode.

\subsection{Experimental Subjects}\label{sec:subjects}

We use two publicly-available pre-trained DNN-based steering angle prediction 
models, i.e., Autumn~\cite{autumn} and Chauffeur~\cite{chauffeur},
that have been widely used in previous work to evaluate various DNN testing 
approaches~\cite{DeepTest,DeepRoad,Surprise}.

Autumn consists of an image preprocessing module implemented using OpenCV to 
compute the optical flow of raw images, and a Convolutional Neural Network (CNN) 
implemented using Tensorflow and Keras to predict steering angles. Chauffeur 
consists of one CNN that extracts the features of input images and a Recurrent 
Neural Network (RNN) that predicts steering angles from the previous 100 consecutive 
images with the aid of a LSTM (Long Short-Term Memory) module. Chauffeur is also 
implemented by Tensorflow and Keras.

\begin{figure}
	\centering
	\includegraphics[width=\linewidth]{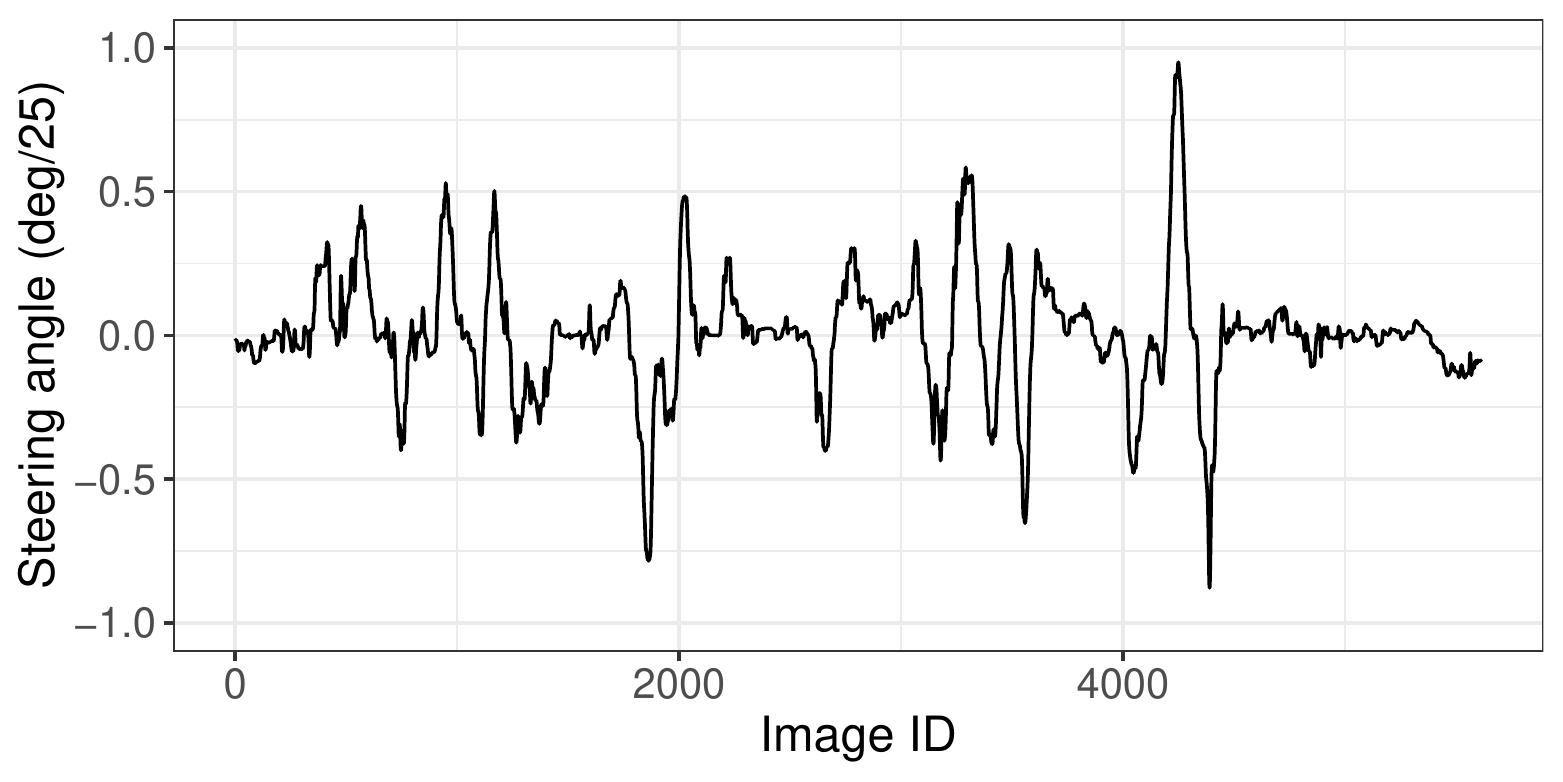}
	\caption{Actual steering angles for the 5614 real-world images for testing}
	\label{fig:testing-data-steering}
\end{figure}

The models are developed using the Udacity dataset~\cite{udacity:dataset}, which 
contains 33808 images for training and 5614 images for testing. The images are  
sequences of frames of two separate videos, one for training and one for testing,  
recorded by a dashboard camera with 20 Frame-Per-Second (FPS). The dataset also 
provides, for each image, the actual steering angle produced by a human driver 
while the videos were recorded. A positive (+) steering angle represents turning 
right, a negative (-) steering angle represents turning left, and a zero angle 
represents staying on a straight line.
The steering angle values are normalized (i.e., they are between $-1$ and 
$+1$) where a $+1$ steering angle value indicates \ang{+25}, and a $-1$ steering 
angle value indicates \ang{-25}\footnote{This is how Tian et 
al.~\cite{DeepTest} have interpreted the steering angle values provided along 
with the Udacity dataset, and we follow their interpretation. We were not able 
to find any explicit information about the measurement unit of these values 
anywhere else.}. Figure~\ref{fig:testing-data-steering} shows the actual steering 
angle values for the sequence of 5614 images in the test dataset. We note that 
the order of images in the training and test datasets matters and is accounted 
for when applying the DNN models. As shown in the figure, the 
steering angles issued by the driver vary considerably over time. The large 
steering angle values (more than \ang{3}) indicate actual road curves,  
while the smaller fluctuations are due to the natural behavior of the human 
driver even when the car drives on a straight road.

Table~\ref{table:rmse} shows the RMSE and MAE values obtained by applying the 
two models to the Udacity test dataset. Note that we were not able to exactly 
replicate the RMSE value reported on the Udacity self-driving challenge 
website~\cite{udacity:challenge}, as the 
value in Table~\ref{table:rmse} is slightly different from those provided by 
Udacity. Reproducibility is known to be a challenge for state-of-the-art deep 
learning methods~\cite{keynote:Joelle} since they involve many parameters 
and details whose variations may lead to different results. To 
enable replication of our work, we have made our detailed configurations (e.g., 
python and auxiliary library versions), together with  supporting materials, 
available online~\cite{supp}.

\begin{table}
\centering
\caption{Details of the subject DNN-based models}
\label{table:rmse}
\begin{tabular}{lrrr}
\toprule
Model & Reported RMSE & Our RMSE & Our MAE  \\
\midrule
Autumn		&	 Not Presented & 0.049 & 0.034	\\
Chauffeur	&	 0.058 & 0.092 & 0.055	\\
\bottomrule
\end{tabular}
\end{table}

While MAE and RMSE are two of the most common metrics used to measure prediction 
errors for learning models with continuous variable outputs, we mainly use MAE 
throughout this paper because, in contrast to RMSE, the MAE values can be 
directly compared with individual steering angle values. For example, 
$\mathit{MAE}(d, \mathbf{r}) = 1$ means that the average prediction error of $d$ 
for the images in $\mathbf{r}$ is $1$ (\ang{25}). Since MAE is a more intuitive metric 
for our purpose, we will only report MAE values in the remainder of our paper.

\subsection{RQ0: Comparing Offline Testing Results for Real-life Data and Simulator-generated Data}

\subsubsection{Setup}
We aim to generate simulator-generated datasets closely mimicking the Udacity 
real-life test dataset and verify whether the prediction errors obtained by applying 
DNNs to the simulator-generate datasets are comparable with those obtained for 
their corresponding real-life ones. As explained in 
Section~\ref{sec:subjects}, our real-life test dataset is a sequence of 5614 
images labelled by their corresponding actual steering angles.
If we could precisely extract the properties of the environment and the dynamics 
of the ego vehicle from the real-life datasets in terms of initial configuration 
parameters of the simulator, we could perhaps generate simulated data resembling 
the real-life videos with high accuracy. However, extracting information from 
real-life video images in a way that the information can be used as inputs of a 
simulator is not possible.

Instead, we propose a two-step heuristic approach to replicate the real-life 
dataset using our simulator. Basically, we steer the simulator to generate a 
sequence of images similar to the images in the real-life dataset such that the 
steering angles generated by the simulator are also close to the steering angle 
labels in the real-life dataset.

In the first step, we observe the test dataset and manually identify the 
information in the images that correspond to some configurable parameter values 
in our domain model described in Section~\ref{sec:domain}. We then create a 
restricted domain model by fixing the parameters in our domain model to the 
values we identified by observing the images in the Udacity test dataset. This 
enables us to steer the simulator to resemble the characteristics of the images 
in the test dataset to the extent possible. Our restricted domain model includes 
the entities and attributes that are neither gray-colored nor bold in 
Figure~\ref{fig:dm}. For example, the restricted domain model does not include 
weather conditions other than sunny because the test dataset has only sunny 
images. This guarantees that the simulator-generated images based on the 
restricted domain model represent sunny scenes only. Using the restricted domain 
model, we randomly generate a large number of scenarios yielding a large number 
of simulator-generated datasets.

In the second step, we aim to ensure that the datasets generated by the 
simulator have similar steering angle labels as the labels in the real-life 
dataset. To ensure this, we match the simulator-generated datasets with 
(sub)sequences of the Udacity test dataset such that the similarities between 
their steering angles are maximized. Note that steering angle is \emph{not} a 
configurable variable in our domain model, and hence, we could not force the 
simulator to generate data with specific steering angle values as those in the 
test dataset by restricting our domain model. Hence, we minimize the 
differences by selecting the closest simulator-generated datasets from a large 
pool of randomly generated ones. To do this, we define, below, the notion of 
``comparability'' between a real-life dataset and a simulator-generated dataset 
in terms of steering angles.

Let $S$ be a set of randomly generated scenarios using the restricted domain 
model, and let $\mathbf{r} = \langle (i^r_1, \theta^r_1), \dots, (i^r_k, 
\theta^r_k) \rangle$ be the Udacity test dataset where $k= 5614$. We denote by 
$\mathbf{r}_{(x,l)} = \langle (i^r_{x+1}, \theta^r_{x+1}), \dots, (i^r_{x+l}, 
\theta^r_{x+l})\rangle$ a subsequence of $\mathbf{r}$ with length $l$ starting 
from index $x+1$ where $x\in \{0,\ldots,k\}$. For a given simulator-generated 
dataset $\mathit{sim}(s) = \langle (i^s_1, \theta^s_1), \dots, (i^s_n, 
\theta^s_n) \rangle$ corresponding to a scenario $s\in S$, we compute 
$\mathbf{r}_{(x,l)}$ using the following three conditions:

\begin{align}
l &= n \label{eq:1} \\
x &= \argmin_{x} \sum_{j=1}^{l} |\theta^s_j - \theta^r_{x+j}| \label{eq:2} \\
&\frac{\sum_{j=1}^{l} |\theta^s_j - \theta^r_{x+j}|}{l} \le \epsilon \label{eq:3}
\end{align}
where $\argmin_x f(x)$ returns\footnote{If $f$ has multiple points of the 
minima, one of them is randomly returned.} $x$ minimizing $f(x)$, and $\epsilon$ 
is a small threshold on the average steering angle difference between 
$\mathit{sim}(s)$ and $\mathbf{r}_{(x,l)}$. 
We say datasets $\mathit{sim}(s)$ and $\mathbf{r}_{(x,l)}$ are \emph{comparable} 
if and only if $\mathbf{r}_{(x,l)}$ satisfies the three above conditions (i.e., 
\ref{eq:1}, \ref{eq:2} and \ref{eq:3}). 

Given the above formalization, our approach to replicate the real-life dataset 
$\mathbf{r}$ using our simulator can be summarized as follows: 
In the first step, we randomly generate a set of many scenarios $S$ based on the 
reduced domain model. In the second step, for every scenario $s \in S$, we 
identify a subsequence  $\mathbf{r}_{(x,l)} | \mathbf{r}$ such that 
$\mathit{sim}(s)$ and $\mathbf{r}_{(x,l)}$ are comparable. 

If $\epsilon$ is too large, we may find $\mathbf{r}_{(x,l)}$ whose steering 
angles are too different from those in $\mathit{sim}(s)$. On the other hand, if 
$\epsilon$ is too small, we may not able to find $\mathbf{r}_{(x,l)}$ that is 
comparable to $\mathit{sim}(s)$ for many randomly generated scenarios $s \in S$ 
in the first step. In our experiments, we select $\epsilon = 0.1$ (\ang{2.5}) 
since, based on our preliminary evaluations, we can achieve an optimal balance 
with this threshold. 

For each comparable pair $\mathit{sim}(s)$ and $\mathbf{r}_{(x,l)}$, we measure 
and compare the prediction errors, i.e., $\mathit{MAE}(d, \mathit{sim}(s))$ and 
$\mathit{MAE}(d, \mathbf{r}_{(x,l)})$ of a DNN $d$. Recall that offline testing 
results for a given DNN $d$ are measured based on prediction errors in terms of 
MAE. If $|\mathit{MAE}(d, \mathit{sim}(s)) - \mathit{MAE}(d, 
\mathbf{r}_{(x,l)})| \le 0.1$ (meaning \ang{2.5} of average prediction error 
across all images), we say that $\mathbf{r}_{(x,l)}$ and $\mathit{sim}(s)$ 
yields \emph{consistent} offline testing results for $d$.

\subsubsection{Results}\label{sec:rq0-results}
\begin{figure}
     \centering
     \begin{subfigure}[b]{\linewidth}
         \centering
         \includegraphics[width=0.95\linewidth]{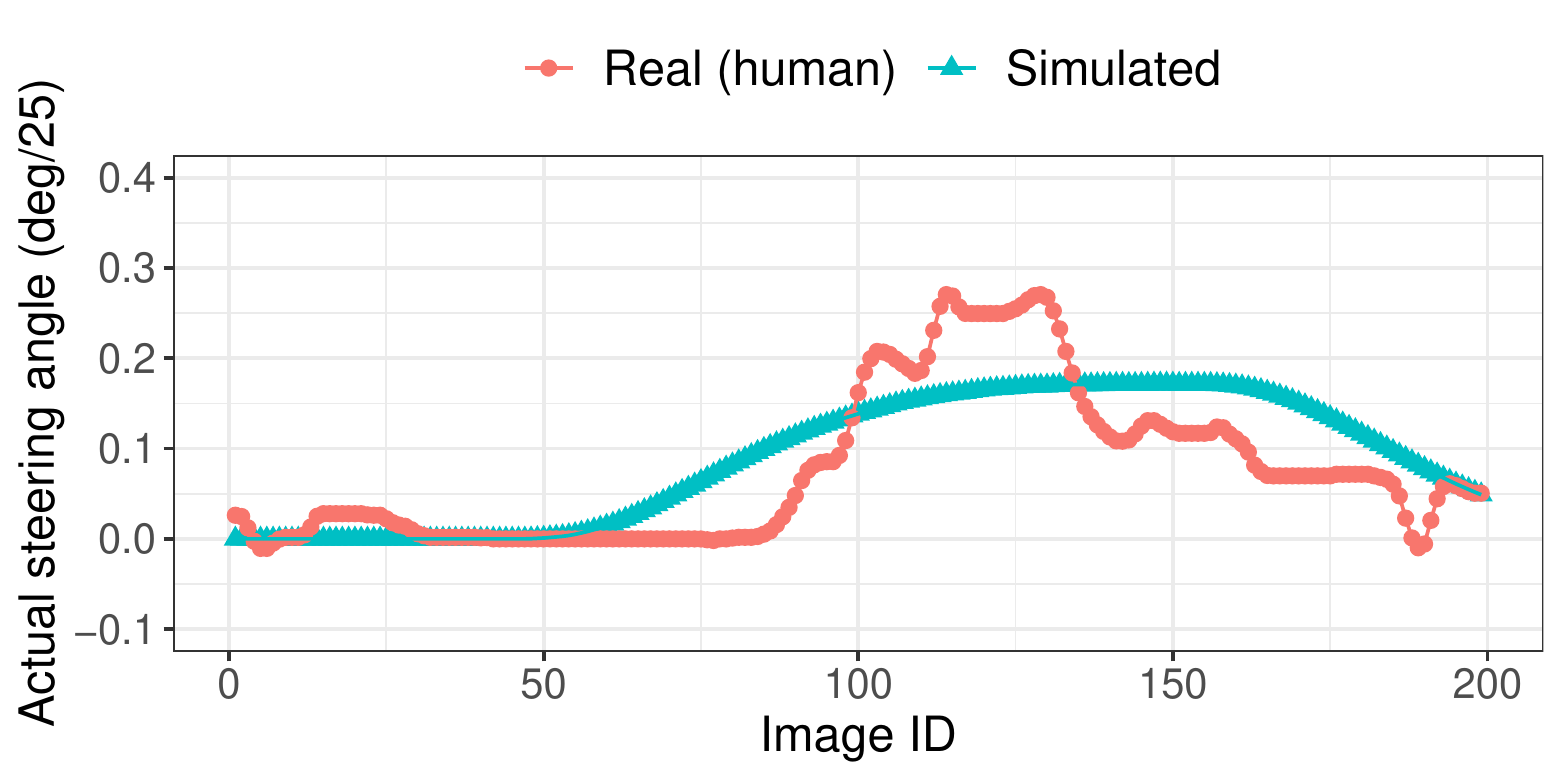}
         \caption{Actual steering angles}
         \label{fig:comparable-steering}
     \end{subfigure}
	\\[3ex]
     \begin{subfigure}[b]{0.48\linewidth}
         \centering
         \includegraphics[width=\linewidth]{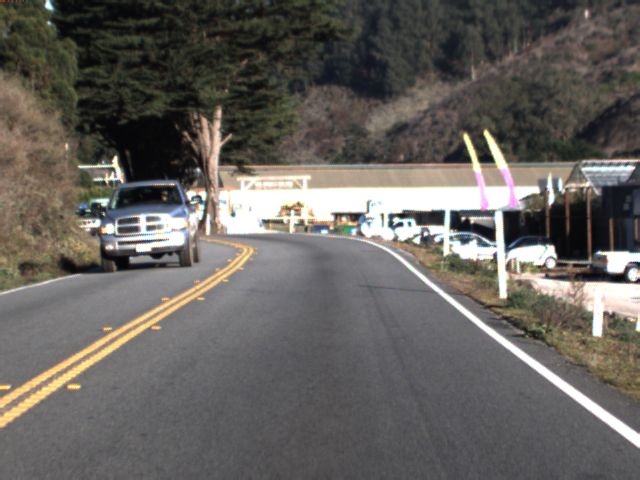}
         \caption{The 64th real image}
         \label{fig:comparable-real}
     \end{subfigure}
     \hfill
     \begin{subfigure}[b]{0.48\linewidth}
         \centering
         \includegraphics[width=\linewidth]{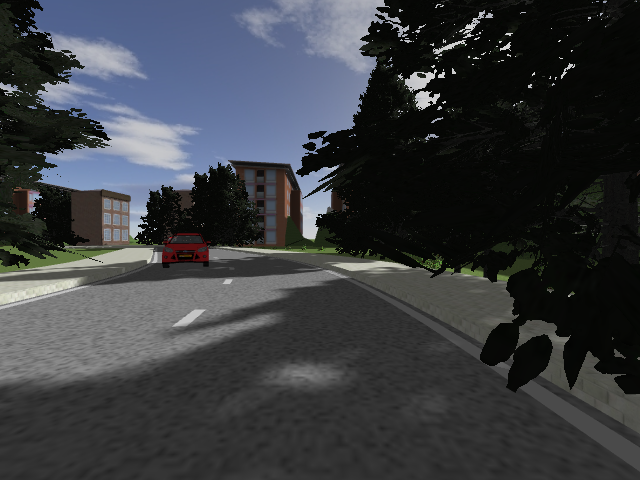}
         \caption{The 64th simulated image}
         \label{fig:comparable-simulated}
     \end{subfigure}
        \caption{Example comparable pair of a simulator-generated 
	dataset and a real-life dataset subsequence}
        \label{fig:comparable-pair}
\end{figure}

Among the 100 randomly generated scenarios (i.e., $|S| = 100$), we identified 92 
scenarios that could match subsequences of the Udacity real-life test dataset. 
Figure~\ref{fig:comparable-steering} shows the steering angles for an example 
comparable pair $\mathit{sim}(s)$ and $\mathbf{r}_{(x,l)}$ in our experiment, 
and Figure~\ref{fig:comparable-real} and \ref{fig:comparable-simulated} show two 
example matching frames from $\mathbf{r}_{(x,l)}$ (i.e., real dataset) and 
$\mathit{sim}(s)$ (i.e., simulator-generated dataset), respectively. 
As shown in the steering angle graph in Figure~\ref{fig:comparable-steering}, 
the simulator-generated dataset and its comparable real dataset subsequence have 
very similar steering angles. Note that the actual steering angles issued by a 
human driver have natural fluctuations while the steering angles generated by 
the simulator are very smooth. Also, the example matching images in 
Figure~\ref{fig:comparable-real} and \ref{fig:comparable-simulated} look quite similar.

\begin{figure}
	\centering
	\includegraphics[width=0.7\linewidth]{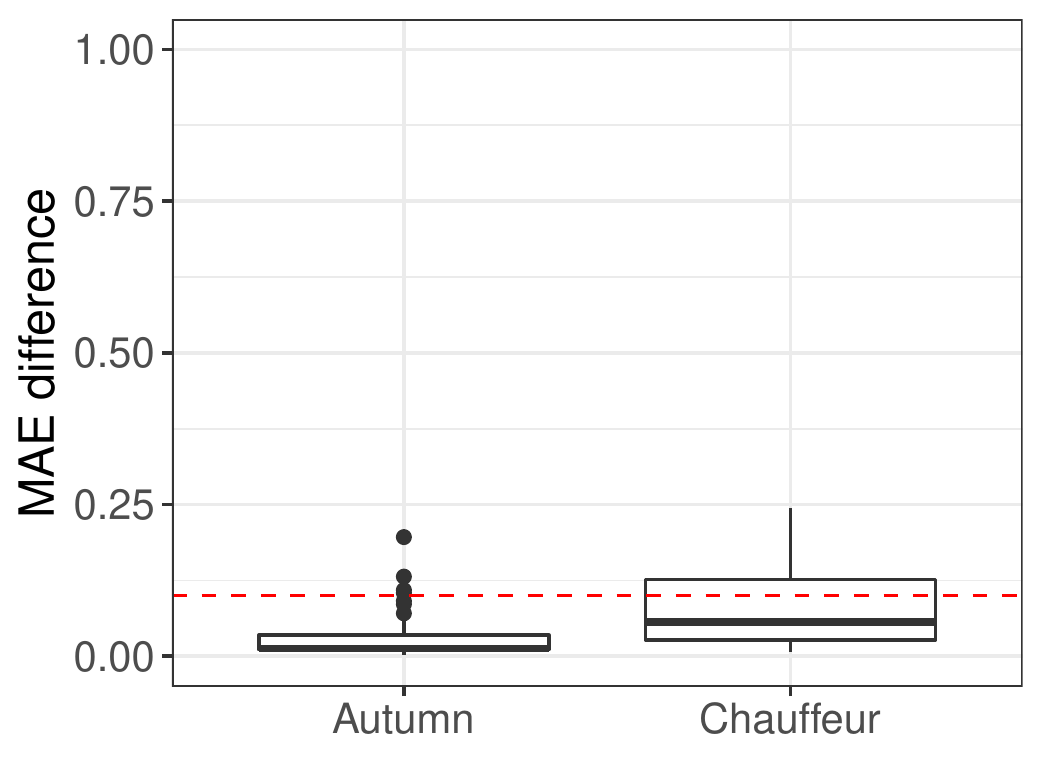}
	\caption{Distributions of the differences between the prediction errors 
	obtained for the real datasets (subsequences) and the simulator-generated datasets}
	\label{fig:rq1-boxplot}
\end{figure}

Figure~\ref{fig:rq1-boxplot} shows the distributions of the differences between 
the prediction errors obtained for the real datasets (subsequences) and the 
simulator-generated datasets for each of our DNNs, Autumn and Chauffeur. 
For Autumn, the average prediction error difference for the real datasets and 
the simulator-generated datasets is $0.027$. Further, 96.7\% of the comparable 
pairs show a prediction error difference below 0.1 (\ang{2.5}). This means that 
the (offline) testing results obtained for the simulator-generated datasets are 
consistent with those obtained using the real-world datasets for almost all 
comparable dataset pairs.
On the other hand, for Chauffeur, 68.5\% of the comparable pairs show a 
prediction error difference below $0.1$. This means that the testing results 
between the real datasets and the simulator-generated datasets are inconsistent 
in 31.5\% of the 92 comparable pairs. Specifically, for \emph{all} of the 
inconsistent case, we observed that  the MAE value for the simulator-generated 
dataset is greater than the MAE value for the real-world dataset. It is 
therefore clear that the prediction error of Chauffeur tends to be larger for 
the simulator-generated dataset than the real-world dataset. In other words, the 
simulator-generated datasets tend to be conservative for Chauffeur and report 
more false positives than for Autumn in terms of prediction errors. We also 
found that, in several cases, Chauffeur's prediction errors are greater than 
$0.2$ while Autumn's prediction errors are less than $0.1$ for the same 
simulator-generated dataset. One possible explanation is that Chauffeur is 
over-fitted to the texture of real images, while Autumn is not thanks to the 
image preprocessing module.
Nevertheless, the average prediction error differences between the real datasets 
and the simulator-generated datasets is $0.079$ for Chauffeur, which is still 
less than $0.1$. This implies that, although Chauffeur will lead to more false 
positives (incorrect safety violations) than Autumn, the number of false 
positives is still unlikely to be overwhelming.

We remark that the choice of  simulator as well as the way we generate data 
using our selected simulator, based on carefully designed experiments such as 
the ones presented here, are of great importance. Selecting a suboptimal 
simulator may lead to many false positives (i.e., incorrectly identified 
prediction errors) rendering simulator-generated datasets ineffective. 

\begin{framed}
The answer to RQ0 is that the prediction error differences between 
simulator-generated datasets and real-life datasets is less than 0.1, on 
average, for both Autumn and Chauffeur. We conclude that we can use 
simulator-generated datasets as a reliable alternative to real-world datasets 
for testing DNNs.
\end{framed}

\subsection{RQ1: Comparison between Offline and Online Testing Results}

\subsubsection{Setup}
We aim to compare offline and online testing results in this research question. 
We randomly generate 50 scenarios and compare the offline and online testing 
results for each of the simulator-generated datasets.

For the scenario generation, we use the extended domain model (see 
Figure~\ref{fig:dm}) to take advantage of all the feasible features provided by 
the simulator. Specifically, in Figure~\ref{fig:dm}, the gray-colored entities 
and attributes in bold are additionally included in the extended domain model 
compared to the restricted domain model used for RQ0. For example, the (full) 
domain model contains various weather conditions, such as rain, snow, and fog, 
in addition to sunny.

Let $S'$ be the set of randomly generated scenarios based on the (full) domain model.
For each scenario $s\in S'$, we prepare the simulator-generated dataset 
$\mathit{sim}(s)$ for offline testing and measure $\mathit{MAE}(d, 
\mathit{sim}(s))$. For online testing, we measure $\mathit{MDCL}(d, s)$. 

Since MAE and MDCL are different metrics, we cannot directly compare MAE and 
MDCL values. To determine whether the offline and online testing results are 
consistent or not, we set threshold values for MAE and MDCL. If $\mathit{MAE}(d, 
\mathit{sim}(s)) < 0.1$ (meaning the average prediction error is less than 
\ang{2.5}) then we interpret the offline testing result of $d$ for $s$ as 
acceptable. On the other hand, if $\mathit{MDCL}(d, s) < 0.7$ (meaning that the 
departure from the centre of the lane observed during the simulation of $s$ is 
less than around one meter), then we interpret the online testing result of $d$ 
for $s$ as acceptable. If both offline and online testing results of $d$ are 
consistently (un)acceptable, we say that offline and online testing are \emph{in 
agreement} regarding testing $d$ for $s$.

\subsubsection{Results}

\begin{figure}
	\centering
	\includegraphics[width=\linewidth]{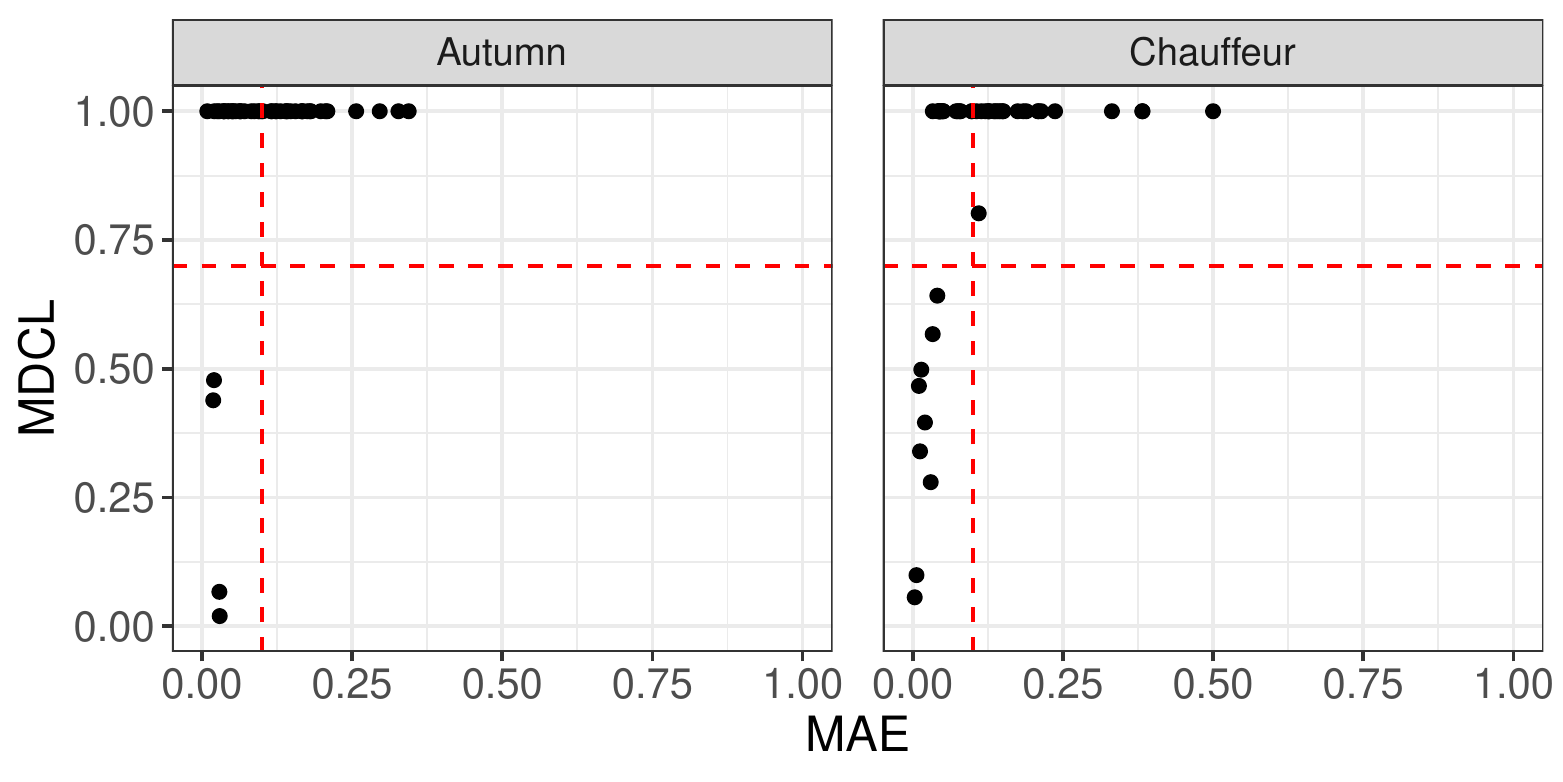}
	\caption{Comparison between offline and online testing results for all scenarios}
	\label{fig:rq2-scatter}
\end{figure}

Figure~\ref{fig:rq2-scatter} shows the comparison between offline and online
testing results in terms of MAE and MDCL values for all the randomly generated 
scenarios in $S'$ where $|S'| = 50$. The x-axis 
is MAE (offline testing) and the y-axis is MDCL (online testing). The dashed 
lines represent the thresholds, i.e., $0.1$ for MAE and $0.7$ for MDCL. 
Table~\ref{table:contingency} provides the number of scenarios classified by the 
offline and online testing results based on the thresholds.
The results show that offline testing and online testing are not in agreement for
44\% and 34\% of the 50 randomly generated scenarios for Autumn and Chauffeur, 
respectively. Surprisingly, offline testing is always more optimistic than online 
testing for the disagreement scenarios. In other words, there is no case where 
the online testing result is acceptable while the offline testing result is not.

\begin{table}
\caption{Number of scenarios classified by \\ offline and online testing results}
\label{table:contingency}
\renewcommand{\arraystretch}{1.3}
    \begin{subtable}{\linewidth}
      \centering
        \caption{Autumn}
\begin{tabular}{|r|rr|r|}
\hline
   & MAE $<$ 0.1 & MAE $\ge$ 0.1 & Total\\ \hline
MDCL $<$ 0.7       & 4                 & 0         & 4              \\
MDCL $\ge$ 0.7 & 22                & 24          & 46            \\ \hline
Total & 26 & 24 & 50 \\ \hline
\end{tabular}
    \end{subtable}%
    \vspace*{7pt}
    \begin{subtable}{\linewidth}
      \centering

        \caption{Chauffeur}
\begin{tabular}{|r|rr|r|}
\hline
    & MAE $<$ 0.1 & MAE $\ge$ 0.1 & Total \\ \hline
MDCL $<$ 0.7       & 9                 & 0           & 9            \\
MDCL $\ge$ 0.7 & 17                & 24            & 41          \\ \hline
Total & 26 & 24 & 50 \\ \hline
\end{tabular}
    \end{subtable}
\end{table}

\begin{figure}
     \centering
     \begin{subfigure}[b]{0.49\linewidth}
         \centering
         \includegraphics[width=\linewidth]{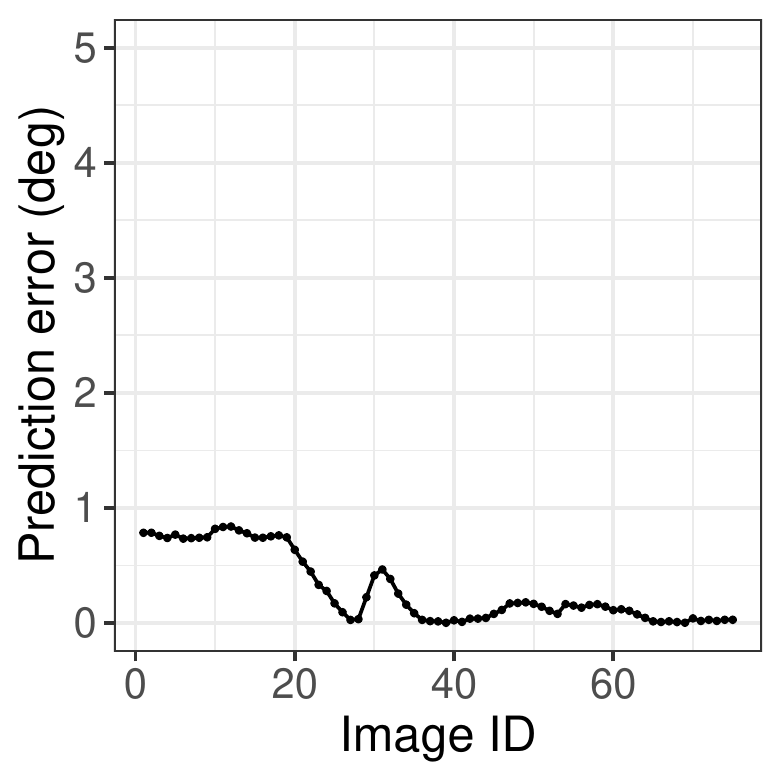}
         \caption{Offline testing result}
         \label{fig:disagree-offline}
     \end{subfigure}
%     \hfill
     \begin{subfigure}[b]{0.49\linewidth}
         \centering
         \includegraphics[width=\linewidth]{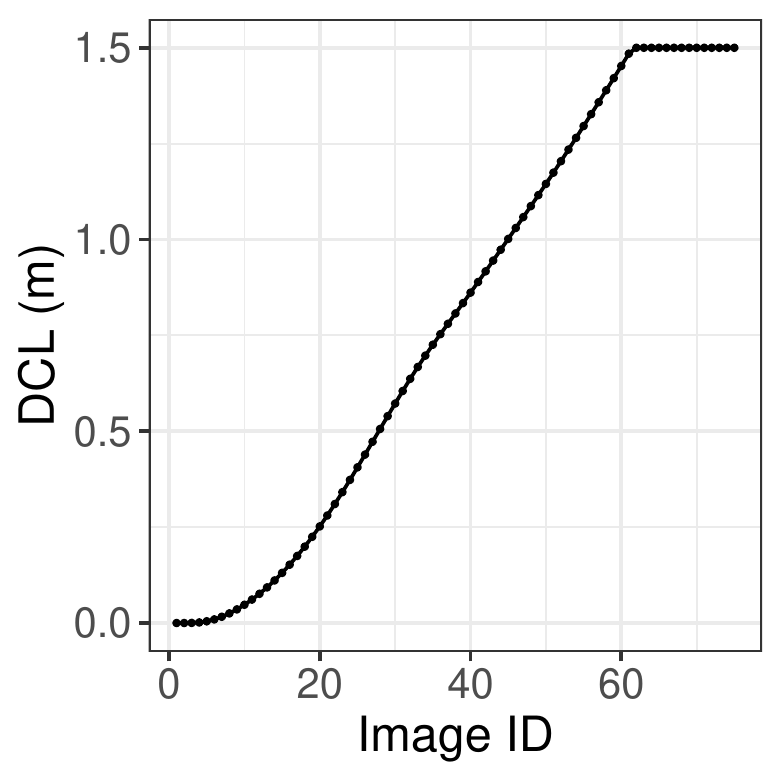}
         \caption{Online testing result}
         \label{fig:disagree-online}
     \end{subfigure}
        \caption{Example disagreed results between offline and online testing}
        \label{fig:rq2-disagree}
\end{figure}

Figure~\ref{fig:rq2-disagree} shows one of the scenarios on which 
offline and online testing disagreed. As shown in Figure~\ref{fig:disagree-offline}, 
the prediction error of the DNN for each image is always less than \ang{1}. 
This means that the DNN appears to be accurate enough according to offline testing. 
However, based on the online testing result in Figure~\ref{fig:disagree-online}, 
the ego vehicle departs from the center of the lane in a critical way 
(i.e., more that \SI{1.5}{\meter}). This is because, over 
time, small prediction errors accumulate, eventually causing a critical lane 
departure. Such accumulation of errors over time is only observable in 
online testing, and this also explains why there is no case where the online 
testing result is acceptable while the offline testing result is not.

The experimental results imply that offline testing cannot properly reveal  
safety violations in ADS-DNNs, because it does not consider their closed-loop 
behavior. Having very acceptably small errors on single images does not guarantee
that there will be no safety violations in the driving environment. Considering 
the fact that detecting safety violations in ADS is the ultimate goal of ADS-DNN 
testing, we conclude that online testing is preferable to offline testing for ADS-DNNs.

\begin{framed}
The answer to RQ1 is that offline and online testing results differ in many cases. 
Offline testing is more optimistic than online testing because
the accumulation of errors is not observed in offline testing.
\end{framed}

\subsection{Threats to Validity}\label{sec:threads}
We propose a two-step approach that builds simulator-generated datasets 
comparable to a given real-life dataset. While it achieves its objective, as 
shown in Section~\ref{sec:rq0-results}, the simulated images are still different 
from the real images. However, we confirmed that the prediction errors obtained 
by applying our subject DNNs to the simulator-generated datasets are comparable 
with those obtained for their corresponding real-life datasets. Thus, the 
experimental results that offline and online testing results often disagree with 
each other are valid.

We used a few thresholds that may change the experimental results 
quantitatively. To reduce the chances of misinterpreting the results, we 
selected intuitive and physically interpretable metrics directly to evaluate 
both offline and online test results (i.e, prediction errors and safety 
violations), and defined threshold values based on common sense and experience. 
Further, adopting different threshold values, as long as they are within a 
reasonable range, does not change our findings. For example, if we use 
$\mathit{MAE(d, \mathit{sim}(s))} < 0.05$ as a threshold in offline testing 
results instead of $\mathit{MAE(d, \mathit{sim}(s))} < 0.1$, the numbers of 
scenarios in Table~\ref{table:contingency} change. However, it does not change 
the fact that we have many scenarios for which offline and online testing 
results disagree, nor does it change the conclusion that offline testing is more 
optimistic than online testing. 

Though we focused, in our case study, on lane-keeping DNNs (steering 
prediction)---which have rather simple structures and do not support braking or 
acceleration, our findings are applicable to all DNNs in the context of ADS as 
long as the closed-loop behavior of ADS matters.

%% file: related-work.tex
\section{Related Work}\label{sec:offline-online}
Table~\ref{table:related-papers} summarizes DNN testing approaches specifically 
proposed in the context of autonomous driving systems. Approaches to the  
general problem of testing machine learning systems are discussed in the recent 
survey by Zhang et al.~\cite{zhang2019machine}.

\begin{table*}
\centering
\caption{Summary of DNN testing studies in the context of autonomous driving}
\label{table:related-papers}
\renewcommand{\arraystretch}{1.3}
%\rowcolors{1}{white}{lightgray}
\begin{tabular}{lllp{2.7cm}p{8cm}}
\toprule
Author(s) & Year & Testing mode & DNN's role & Summary \\
\midrule
%\midrule
Dreossi et al.~\cite{dreossi2017systematic} & 2017 & Offline & Object detection & Test image generation by arranging basic objects using greedy search  \\
Pei et al.~\cite{DeepXplore} & 2017 & Offline & Lane keeping & Coverage-based label-preserving test image generation using joint optimization with gradient ascent \\
Tian et al.~\cite{DeepTest} & 2018 & Offline & Lane keeping & Coverage-based label-preserving test image generation using greedy search with simple image transformations \\
\rowcolor{lightgray}
Tuncali et al.~\cite{8500421} & 2018 & Online & Object detection & Test scenario generation using the combination of covering arrays and simulated annealing \\
Wicker et al.~\cite{89960-2_22} & 2018 & Offline & Traffic sign recognition & Adversarial image generation using feature extraction \\
Zhang et al.~\cite{DeepRoad} & 2018 & Offline & Lane keeping & Label-preserving test image generation using Generative Adversarial Networks (GANs) \\
Zhou et al.~\cite{zhou2018deepbillboard} & 2018 & Offline & Lane keeping & Adversarial billboard-image generation for digital and physical adversarial perturbation \\
\rowcolor{lightgray}
Gambi et al.~\cite{Gambi2019issta} & 2019 & Online & Lane keeping & Automatic virtual road network generation using search-based Procedural Content Generation (PCG) \\
Kim et al.~\cite{Surprise} & 2019 & Offline & Lane keeping & Improving the accuracy of DNNs against adversarial examples using surprise adequacy \\
\rowcolor{lightgray}
Majumdar et al.~\cite{majumdar2019paracosm} & 2019 & Online & Object detection, \newline lane keeping & Test scenario description language and simulation-based test scenario 
generation to cover parameterized environments \\
Zhou et al.~\cite{Zhou:2019:MTD} & 2019 & Offline & Object detection & Combination of Metamorphic Testing (MT) and fuzzing for 3-dimensional point cloud data \\
\midrule
\rowcolor{lightgray}
This paper & 2019 & Offline and online & Lane keeping & Comparison between offline and online testing results \\
\bottomrule
\end{tabular}
\end{table*}

In Table~\ref{table:related-papers},  online testing approaches are highlighted 
in grey. As indicated in Table~\ref{table:online-offline},  offline testing 
approaches focus on DNNs as individual units without accounting for the 
closed-loop behavior of a DNN-based ADS.
Most of them aim to generate test data (either images or 3-dimensional point 
clouds) that lead to DNN prediction errors. 
Dreossi et al.~\cite{dreossi2017systematic} synthesized images for driving 
scenes by arranging basic objects (e.g., road backgrounds and vehicles) and 
tuning image parameters (e.g., brightness, contrast, and saturation).
Pei et al.~\cite{DeepXplore} proposed \textsc{DeepXplore}, an approach that 
synthesizes images by solving a joint optimization problem that maximizes both 
neuron coverage (i.e., the rate of activated neurons) and differential behaviors 
of multiple DNNs for the synthesized images.
Tian et al.~\cite{DeepTest} presented \textsc{DeepTest}, an approach that 
generates label-preserving images from training data using greedy search for 
combining simple image transformations (e.g., rotate, scale, and for and rain 
effects) to increase neuron coverage. 
Wicker et al.~\cite{89960-2_22} generated adversarial examples, i.e., small 
perturbations that are almost imperceptible by humans but causing DNN 
misclassifications, using feature extraction from images.
Zhang et al.~\cite{DeepRoad} presented \textsc{DeepRoad}, an approach that 
produces various driving scenes and weather conditions by applying Generative 
Adversarial Networks (GANs) along with corresponding real-world weather scenes. 
Zhou et al.~\cite{Zhou:2019:MTD} combined Metamorphic Testing (MT) and Fuzzing 
for 3-dimensional point cloud data generated by a LiDAR sensor to reveal 
erroneous behaviors of an object detection DNN.
Zhou et al.~\cite{zhou2018deepbillboard} proposed \textsc{DeepBillboard}, an 
approach that produces both digital and physical adversarial billboard images to 
continuously mislead the DNN across dashboard camera frames. While this work is 
different from the other offline testing studies as it introduces adversarial 
attacks through sequences of frames, its goal was still the generation of test 
images to reveal DNN prediction errors. 
In contrast, Kim et al.~\cite{Surprise} defined a coverage criterion, called 
\emph{surprise adequacy}, based on the behavior of DNN-based systems with 
respect to their training data. Images generated by \textsc{DeepTest} were 
sampled to improve such coverage and used to increase the accuracy of the DNN 
against adversarial examples.

Online testing studies exercise the ADS closed-loop behavior and generate test 
driving scenarios that cause safety violations, such as unintended lane 
departure or collision with pedestrians.
Tuncali et al.~\cite{8500421} were the first to raise the problem that previous 
works mostly focused on the DNNs, without accounting for the closed-loop 
behavior of the system.
Gambi et al.~\cite{Gambi2019issta} also pointed out that testing DNNs for ADS 
using only single frames cannot be used to evaluate closed-loop properties of 
ADS. They presented \textsc{AsFault}, a tool that generates virtual roads which 
cause self-driving cars to depart from their lane.
Majumdar et al.~\cite{majumdar2019paracosm} presented a language for describing 
test driving scenarios in a parametric way and provided \textsc{Paracosm}, a 
simulation-based testing tool that generates a set of test parameters in such a 
way as to achieve diversity.
We should note that all the online testing studies rely on virtual (simulated) 
environments, since, as mentioned before, testing DNNs for ADS in real traffic 
is very dangerous and expensive. Further, there is a growing body of evidence 
demonstrating that simulation-based testing is effective at finding violations. 
For example, recent studies for robotic applications show that simulation-based 
testing of robot function models not only reveals most bugs identified  during 
outdoor robot testing, but that it can additionally reveal several bugs that 
could not have been detected by outdoor testing~\cite{8009918}.

In summary, even though online testing has received more attention recently, 
most existing approaches to testing DNN in the context of ADS focus on  offline 
testing. We note that  none of the existing techniques compare offline and 
online testing results, and neither do they demonstrate relative effectiveness 
of test datasets obtained from simulators compared to those captured from 
real-life.

%% file: conclusion.tex
\section{Conclusion}\label{sec:conclusion}

In this paper, we distinguish two general modes of testing, namely offline testing 
and online testing, for DNNs developed in the context of Advanced Driving Systems (ADS). 
Offline testing search for DNN prediction errors based on test datasets obtained independently 
from the DNNs under test, while online testing focuses on detecting safety violations of 
a DNN-based ADS in a closed-loop mode by testing it in interaction with its 
real or simulated application environment. 
Offline testing is less expensive and faster than online testing but may not be effective 
at finding significant errors in DNNs. Online testing is more easily performed and safer 
with a simulator but we have no guarantees that the results are representative of real 
driving environments. 

To address the above concerns, we conducted a case study to compare 
the offline and online testing of DNNs for the end-to-end control of a vehicle. 
We also investigated if we can use simulator-generated datasets as 
a reliable substitute to real-world datasets for DNN testing. The experimental 
results show that simulator-generated datasets yield DNN prediction errors that 
are similar to those obtained by testing DNNs with real-world datasets. Also, 
offline testing appears to be more optimistic than online testing as many safety violations 
identified by online testing were not suggested by offline testing prediction errors. 
Furthermore, large prediction errors generated by offline testing always led to severe safety 
violations detectable by online testing.
Such results have important practical implications for DNN testing in the context of ADS.

As part of future work, we plan to develop an approach that effectively combines both 
offline and online testing to automatically identify critical safety violations. 
We also plan to investigate how to improve the performance of DNN-based ADS 
using the identified prediction errors and safety violations for further learning.

%% file: main.bbl
\begin{thebibliography}{10}
\providecommand{\url}[1]{#1}
\csname url@samestyle\endcsname
\providecommand{\newblock}{\relax}
\providecommand{\bibinfo}[2]{#2}
\providecommand{\BIBentrySTDinterwordspacing}{\spaceskip=0pt\relax}
\providecommand{\BIBentryALTinterwordstretchfactor}{4}
\providecommand{\BIBentryALTinterwordspacing}{\spaceskip=\fontdimen2\font plus
\BIBentryALTinterwordstretchfactor\fontdimen3\font minus
  \fontdimen4\font\relax}
\providecommand{\BIBforeignlanguage}[2]{{%
\expandafter\ifx\csname l@#1\endcsname\relax
\typeout{** WARNING: IEEEtran.bst: No hyphenation pattern has been}%
\typeout{** loaded for the language `#1'. Using the pattern for}%
\typeout{** the default language instead.}%
\else
\language=\csname l@#1\endcsname
\fi
#2}}
\providecommand{\BIBdecl}{\relax}
\BIBdecl

\bibitem{chen2015deepdriving}
C.~Chen, A.~Seff, A.~Kornhauser, and J.~Xiao, ``Deepdriving: Learning
  affordance for direct perception in autonomous driving,'' in
  \emph{Proceedings of the IEEE International Conference on Computer Vision},
  2015, pp. 2722--2730.

\bibitem{bojarski2016end}
M.~Bojarski, D.~Del~Testa, D.~Dworakowski, B.~Firner, B.~Flepp, P.~Goyal, L.~D.
  Jackel, M.~Monfort, U.~Muller, J.~Zhang \emph{et~al.}, ``End to end learning
  for self-driving cars,'' \emph{arXiv preprint arXiv:1604.07316}, 2016.

\bibitem{chi2017deep}
L.~Chi and Y.~Mu, ``Deep steering: Learning end-to-end driving model from
  spatial and temporal visual cues,'' \emph{arXiv preprint arXiv:1708.03798},
  2017.

\bibitem{ciresan2012}
\BIBentryALTinterwordspacing
D.~C. Ciresan, U.~Meier, and J.~Schmidhuber, ``Multi-column deep neural
  networks for image classification,'' \emph{CoRR}, vol. abs/1202.2745, 2012.
  [Online]. Available: \url{http://arxiv.org/abs/1202.2745}
\BIBentrySTDinterwordspacing

\bibitem{SutskeverVL14}
I.~Sutskever, O.~Vinyals, and Q.~V. Le, ``Sequence to sequence learning with
  neural networks,'' in \emph{Advances in Neural Information Processing Systems
  27}, Z.~Ghahramani, M.~Welling, C.~Cortes, N.~D. Lawrence, and K.~Q.
  Weinberger, Eds.\hskip 1em plus 0.5em minus 0.4em\relax Curran Associates,
  Inc., 2014, pp. 3104--3112.

\bibitem{DengHK13}
L.~{Deng}, G.~{Hinton}, and B.~{Kingsbury}, ``New types of deep neural network
  learning for speech recognition and related applications: an overview,'' in
  \emph{2013 IEEE International Conference on Acoustics, Speech and Signal
  Processing}, May 2013, pp. 8599--8603.

\bibitem{DeepXplore}
\BIBentryALTinterwordspacing
K.~Pei, Y.~Cao, J.~Yang, and S.~Jana, ``Deepxplore: Automated whitebox testing
  of deep learning systems,'' in \emph{Proceedings of the 26th Symposium on
  Operating Systems Principles}, ser. SOSP '17.\hskip 1em plus 0.5em minus
  0.4em\relax New York, NY, USA: ACM, 2017, pp. 1--18. [Online]. Available:
  \url{http://doi.acm.org/10.1145/3132747.3132785}
\BIBentrySTDinterwordspacing

\bibitem{DeepTest}
\BIBentryALTinterwordspacing
Y.~Tian, K.~Pei, S.~Jana, and B.~Ray, ``Deeptest: Automated testing of
  deep-neural-network-driven autonomous cars,'' in \emph{Proceedings of the
  40th International Conference on Software Engineering}, ser. ICSE '18.\hskip
  1em plus 0.5em minus 0.4em\relax New York, NY, USA: ACM, 2018, pp. 303--314.
  [Online]. Available: \url{http://doi.acm.org/10.1145/3180155.3180220}
\BIBentrySTDinterwordspacing

\bibitem{DeepRoad}
\BIBentryALTinterwordspacing
M.~Zhang, Y.~Zhang, L.~Zhang, C.~Liu, and S.~Khurshid, ``Deeproad: Gan-based
  metamorphic testing and input validation framework for autonomous driving
  systems,'' in \emph{Proceedings of the 33rd ACM/IEEE International Conference
  on Automated Software Engineering}, ser. ASE 2018.\hskip 1em plus 0.5em minus
  0.4em\relax New York, NY, USA: ACM, 2018, pp. 132--142. [Online]. Available:
  \url{http://doi.acm.org/10.1145/3238147.3238187}
\BIBentrySTDinterwordspacing

\bibitem{DeepGauge}
\BIBentryALTinterwordspacing
L.~Ma, F.~Juefei-Xu, F.~Zhang, J.~Sun, M.~Xue, B.~Li, C.~Chen, T.~Su, L.~Li,
  Y.~Liu, J.~Zhao, and Y.~Wang, ``Deepgauge: Multi-granularity testing criteria
  for deep learning systems,'' in \emph{Proceedings of the 33rd ACM/IEEE
  International Conference on Automated Software Engineering}, ser. ASE
  2018.\hskip 1em plus 0.5em minus 0.4em\relax New York, NY, USA: ACM, 2018,
  pp. 120--131. [Online]. Available:
  \url{http://doi.acm.org/10.1145/3238147.3238202}
\BIBentrySTDinterwordspacing

\bibitem{zhou2018deepbillboard}
H.~Zhou, W.~Li, Y.~Zhu, Y.~Zhang, B.~Yu, L.~Zhang, and C.~Liu, ``Deepbillboard:
  Systematic physical-world testing of autonomous driving systems,'' 2018.

\bibitem{zhang2019machine}
J.~M. Zhang, M.~Harman, L.~Ma, and Y.~Liu, ``Machine learning testing: Survey,
  landscapes and horizons,'' 2019.

\bibitem{udacity:challenge}
``Udacity self-driving challenge 2,''
  \url{https://github.com/udacity/self-driving-car/tree/master/challenges/challenge-2},
  2016, accessed: 2019-10-11.

\bibitem{prescan}
{TASS International - Siemens Group}, ``Prescan: Simulation of adas and active
  safety,'' \url{https://tass.plm.automation.siemens.com}, accessed:
  2019-10-11.

\bibitem{supp}
``Supporting materials (temporal link for the double-blind review),''
  \url{https://tinyurl.com/ICST-2020}, accessed: 2019-10-14.

\bibitem{6248074}
A.~{Geiger}, P.~{Lenz}, and R.~{Urtasun}, ``Are we ready for autonomous
  driving? the kitti vision benchmark suite,'' in \emph{2012 IEEE Conference on
  Computer Vision and Pattern Recognition}, June 2012, pp. 3354--3361.

\bibitem{pomerleau1989alvinn}
D.~A. Pomerleau, ``Alvinn: An autonomous land vehicle in a neural network,'' in
  \emph{Advances in neural information processing systems}, 1989, pp. 305--313.

\bibitem{Surprise}
\BIBentryALTinterwordspacing
J.~Kim, R.~Feldt, and S.~Yoo, ``Guiding deep learning system testing using
  surprise adequacy,'' in \emph{Proceedings of the 41st International
  Conference on Software Engineering}, ser. ICSE '19.\hskip 1em plus 0.5em
  minus 0.4em\relax Piscataway, NJ, USA: IEEE Press, 2019, pp. 1039--1049.
  [Online]. Available: \url{https://doi.org/10.1109/ICSE.2019.00108}
\BIBentrySTDinterwordspacing

\bibitem{prosivic}
{ESI Group}, ``Esi pro-sivic - 3d simulations of environments and sensors,''
  \url{https://github.com/ApolloAuto/apollo}, accessed: 2019-10-11.

\bibitem{carla}
A.~Dosovitskiy, G.~Ros, F.~Codevilla, A.~Lopez, and V.~Koltun, ``{CARLA}: {An}
  open urban driving simulator,'' in \emph{Proceedings of the 1st Annual
  Conference on Robot Learning}, pp. 1--16.

\bibitem{apollo}
{Baidu}, ``Apollo open platform,''
  \url{https://www.esi-group.com/software-solutions/virtual-environment/virtual-systems-controls/esi-pro-sivictm-3d-simulations-environments-and-sensors},
  accessed: 2019-10-11.

\bibitem{udacity:dataset}
``Udacity self-driving challenge 2, ch2-001 (testing) and ch2-002 (training),''
  \url{https://github.com/udacity/self-driving-car/tree/master/datasets/CH2},
  2016, accessed: 2019-10-11.

\bibitem{OCL}
``Object constraint language specification,''
  \url{https://www.omg.org/spec/OCL/}, accessed: 2019-10-11.

\bibitem{autumn}
``Autumn model,''
  \url{https://github.com/udacity/self-driving-car/tree/master/steering-models/community-models/autumn},
  2016, accessed: 2019-10-11.

\bibitem{chauffeur}
``Chauffeur model,''
  \url{https://github.com/udacity/self-driving-car/tree/master/steering-models/community-models/chauffeur},
  2016, accessed: 2019-10-11.

\bibitem{keynote:Joelle}
J.~Pineau, ``Icse 2019 keynote: Building reproducible, reusable, and robust
  machine learning software,''
  \url{https://2019.icse-conferences.org/details/icse-2019-Plenary-Sessions/20/Building-Reproducible-Reusable-and-Robust-Machine-Learning-Software},
  May 2019, accessed: 2019-10-11.

\bibitem{dreossi2017systematic}
T.~Dreossi, S.~Ghosh, A.~Sangiovanni-Vincentelli, and S.~A. Seshia,
  ``Systematic testing of convolutional neural networks for autonomous
  driving,'' 2017.

\bibitem{8500421}
C.~E. {Tuncali}, G.~{Fainekos}, H.~{Ito}, and J.~{Kapinski}, ``Simulation-based
  adversarial test generation for autonomous vehicles with machine learning
  components,'' in \emph{2018 IEEE Intelligent Vehicles Symposium (IV)}, June
  2018, pp. 1555--1562.

\bibitem{89960-2_22}
M.~Wicker, X.~Huang, and M.~Kwiatkowska, ``Feature-guided black-box safety
  testing of deep neural networks,'' in \emph{Tools and Algorithms for the
  Construction and Analysis of Systems}, D.~Beyer and M.~Huisman, Eds.\hskip
  1em plus 0.5em minus 0.4em\relax Cham: Springer International Publishing,
  2018, pp. 408--426.

\bibitem{Gambi2019issta}
\BIBentryALTinterwordspacing
A.~Gambi, M.~Mueller, and G.~Fraser, ``Automatically testing self-driving cars
  with search-based procedural content generation,'' in \emph{Proceedings of
  the 28th ACM SIGSOFT International Symposium on Software Testing and
  Analysis}, ser. ISSTA 2019.\hskip 1em plus 0.5em minus 0.4em\relax New York,
  NY, USA: ACM, 2019, pp. 318--328. [Online]. Available:
  \url{http://doi.acm.org/10.1145/3293882.3330566}
\BIBentrySTDinterwordspacing

\bibitem{majumdar2019paracosm}
R.~Majumdar, A.~Mathur, M.~Pirron, L.~Stegner, and D.~Zufferey, ``Paracosm: A
  language and tool for testing autonomous driving systems,'' 2019.

\bibitem{Zhou:2019:MTD}
\BIBentryALTinterwordspacing
Z.~Q. Zhou and L.~Sun, ``Metamorphic testing of driverless cars,''
  \emph{Commun. ACM}, vol.~62, no.~3, pp. 61--67, Feb. 2019. [Online].
  Available: \url{http://doi.acm.org/10.1145/3241979}
\BIBentrySTDinterwordspacing

\bibitem{8009918}
T.~{Sotiropoulos}, H.~{Waeselynck}, J.~{Guiochet}, and F.~{Ingrand}, ``Can
  robot navigation bugs be found in simulation? an exploratory study,'' in
  \emph{2017 IEEE International Conference on Software Quality, Reliability and
  Security (QRS)}, July 2017, pp. 150--159.

\end{thebibliography}
